\documentclass[final]{IEEEtran}

\usepackage{graphicx}  
\usepackage{amsmath}   

\usepackage{algorithm}
\usepackage{algorithmic}
\usepackage{epsfig}
\usepackage{amssymb}
\usepackage{bm}

\newtheorem{definition}{Definition}

\newcommand{\comment}[1]{}
\newfont{\bboard}{msbm10 scaled\magstephalf}

\begin{document}

\title{ Kernel Two-Sample Hypothesis Testing Using Kernel Set Classification}

\author{Hamed~Masnadi-Shirazi \\
        School of Electrical and Computer Engineering, \\
				Shiraz University, \\
				Shiraz, Iran \\
}

\maketitle

\begin{abstract}
The two-sample hypothesis testing problem is studied for the challenging scenario of high dimensional data sets with small sample sizes.
We  show that the two-sample hypothesis testing problem can be posed as a one-class set classification problem.
In the  set classification problem  the goal is to classify a set of data points that are assumed to have a common class. 
We prove that the average probability of error given a set is less than or equal to the Bayes error and decreases as a power of $n$ number of sample data points in the set. We use the  positive definite Set Kernel for directly mapping sets of data to an  associated Reproducing Kernel Hilbert Space, without the need
to learn a probability distribution.  We specifically solve the two-sample hypothesis testing problem using a one-class SVM in conjunction with the proposed Set Kernel.   We compare the proposed method with the Maximum Mean Discrepancy, F-Test and T-Test methods on a number of challenging simulated high dimensional and small sample size data. We also perform two-sample hypothesis testing experiments on six cancer gene expression data sets and achieve zero type-I and type-II error results on all data sets.  
\end{abstract}

\begin{keywords}
Set Classification, Positive Definite Kernel, Two-Sample Hypothesis Testing, One-Class Classification,  Maximum Mean Discrepancy,  Gene Expression Data
\end{keywords}

\IEEEpeerreviewmaketitle

\section{Introduction}
\label{intro}
Many  problems are naturally  in the form of a set classification problem which is defined as classifying a set of data points given that all the data points in the set belong to the same unknown class \cite{KernelForSets, SetImageNonParamModel, SetClass}. In other words, in a set classification problem we classify a set of data vectors rather than a single vector. For example the pixels of an image can be thought of as a set and classifying the image can be thought of as classifying the set of pixels in the image \cite{KernelForSets}. As another example, face recognition from multiple images of the same person can be posed as a set classification problem where the set of multiple images  must be assigned to a certain individual \cite{FaceSetLongTerm, FaceSet, FaceRecogSetsBased}. Many other problems such as gene expression or chemical classification, document classification, ontology alignment, scene classification, video classification and multiple pose object recognition  can  be naturally posed as set classification problems \cite{SetVideo, ChemSet2, ChemSet, SetClass, SparseImageSet, book:Ontology, ImageSet}. 

There are generally two different approaches to the set classification problem. The first approach basically uses a standard classifier on the individual elements of
the set and then applies a variety of voting schemes to reach a consensus on  the entire set \cite{SetClass}.  In this paper we formally prove in Section \ref{sec:SetRisk} that this approach is suboptimal. Intuitively, these type of methods do not make full use of all the information available and ignore the inter-dependencies between the elements of the set.  They classify each element independently as apposed to using all of the data points concurrently to learn a class for the entire set of samples. 

The second approach is to somehow summarize the set of data points into a single entity and then make a classification decision based on this single entity. 
For example, a simple  approach could be to summarize a set of vectors into an average vector and classify the average vector. A more advanced approach could be to summarize the set of vectors into a specialized probability distribution and then make a decision on the probability distribution
 \cite{KernelForSets, SetParamModel1, SetParamModel3, SetParamModel4, SetFaceParamModel} or to use kernel or nonparametric methods that directly measure distances between sets \cite{SetDistances, SetImageNonParamModel, SetImageNonParamModel2, FaceRecogSetsBased, SparseImageSet}. These methods can suffer from a few deficiencies. Namely, estimating a probability distribution is generally
problematic in high dimensional spaces and many of the proposed kernels are not positive definite. 

We take the second approach while avoiding its pitfalls. Specifically, we use a method where each set of vectors of any size is mapped directly to a vector in a Reproducing Kernel Hilbert Space (RKHS), without the need to model the distribution. Notably, the Set Kernel associated with this RKHS is proven to be a positive definite inner product kernel and the norm associated with this inner product kernel is the 
empirical Maximum Mean Discrepancy (MMD) \cite{TwoMMD, MMD}.  The theoretical properties of the MMD have been extensively studied in \cite{MMD} and the empirical MMD has been justified using performance guarantees.
Rather than looking at the empirical MMD as an approximation to the MMD, it can be independently justified as a norm in a certain RKHS of sets.

The MMD has found many recent applications in diverse fields ranging from image analysis \cite{MMDImage}  and class ratio estimation \cite{MMDClassRatio} to 
nonparametric  scoring rules \cite{MMDScoreRule}. Nevertheless, it was initially introduced as a nonparametric two-sample hypothesis test \cite{MMD}. 
Traditional parametric hypothesis tests such as the T-Test and F-Test \cite{book:TTEST} are not suited for high dimensional data because of poor estimation in high dimensional spaces. The nonparametric MMD method on the other hand, was proven to have the injective property \cite{MMD} and shown to have a significant performance advantage in
high dimensional data problems. For example it was shown to significantly outperform traditional hypothesis testing methods on gene expression data which are high
dimensional in nature \cite{MMD}.

In light of the alternative way of looking at the empirical MMD as a norm or distance measure between sets in an RKHS, the hypothesis testing method based on the empirical MMD can be thought of as a primitive one-class classification problem. Specifically, this hypothesis test is based on learning a threshold for the distances between different sets in the training set. It then tests a set by comparing the distance between it and a training set. The test set is rejected if it is above the threshold.

Here we suggest that the above primitive method can be improved  using more advanced  one-class classifiers \cite{OneClassPhd} given a well defined RKHS for sets. Rather than learn a single threshold on the distances between sets, we  learn a one-class SVM \cite{OneClassSVM} on the RKHS of sets using the associated Set Kernel. The training data sets are used to learn the one-class SVM boundary and a test set is rejected if it falls outside this boundary in the RKHS.

In the experimental section we show that this novel approach to the two-sample hypothesis problem, using the one-class SVM with Set Kernels, leads to state of the art results on challenging simulated and real world data sets. We consider multivariate Gaussian classes with equal means and different variances in various low and high dimensional spaces which are  challenging for both the F-Test and MMD methods. We also consider various gene expression data sets and show that the 
one-class SVM with Set Kernels method has perfect performance on these  high dimensional data sets. 

The paper is organized as follows. In Sections \ref{sec:background1} and \ref{sec:background2} we review some background material on Reproducing Kernel Hilbert Spaces and kernel two-sample hypothesis testing using the MMD. Next, in Section \ref{sec:SetRKHS},  we establish the RKHS for sets by considering the Set Kernel and establishing that it is positive definite.  In Section \ref{sec:SetRisk} we motivate the use of a set classifier by proving that the average probability of error decreases as we use a larger set of sample data points for classification.
The  two-sample hypothesis test method using the one-class SVM with Set Kernels in explained in Section \ref{sec:HypothesisAsSetClassify}. Finally, the two-sample hypothesis test experimental results are presented in Section \ref{sec:Exp} using both simulated data and real world gene expression data.

\section{Background on Reproducing Kernel Hilbert Spaces }
\label{sec:background1}
Data samples in the input space, ${\bf x} \in \cal{X}$,  are typically mapped to a higher dimensional feature space for improved separation between the classes. 
The kernel function can be viewed as an efficient way of computing inner products in this high dimensional feature space \cite{ book:RHKS, Sriperumbudur2010}. 
For a given mapping $\Phi: \cal{X} \rightarrow \cal{H}$, the kernel function $k$ allows us to compute the inner product between two vectors 
$\Phi({\bf x}), \Phi({\bf x}') $ in the feature space
$ \cal{H}$ without having to explicitly know the mapping $\Phi$, in the form of 
\begin{equation}
k({\bf x},{\bf x}')=<\Phi({\bf x}), \Phi({\bf x}') >_{\cal{H}}. 
\end{equation}

If a  function $k$ happens to be  positive definite, meaning that 
\begin{equation}
\sum_{i=1}^{m} \sum_{j=1}^{m} c_i c_j k({\bf x}_i,{\bf x}_j) \ge 0
\end{equation}
for any $m \in \mathbb{N}$, any choice of ${\bf x}_1, {\bf x}_2, ..., {\bf x}_m \in \cal{X}$ and any coefficients $c_1, c_2, ..., c_m \in \mathbb{R}$,  
then there exists a Hilbert space $\cal{H}$
and a mapping $\Phi: \cal{X} \rightarrow \cal{H}$ such that $k$ computes the inner product in that space. In other words, we can write a positive definite function
in the form of an inner product   
$k({\bf x},{\bf x}')=<\Phi({\bf x}), \Phi({\bf x}')>_{\cal{H}}$ and conversely if a function  can be written as an inner product it is positive definite. 

Furthermore, there is a  Reproducing Kernel Hilbert Space (RKHS) associated with every positive definite kernel such that
\begin{equation}
<\Phi({\bf x}), f >_{\cal{H}} = f({\bf x}) \mbox{ for all } f \in \cal{H}
\end{equation}
where the mapping can be written as $\Phi({\bf x})=k({\bf x},.)$ and $k({\bf x},.)$ is the positive definite kernel function parametrized by ${\bf x}$.

\section{Background on Kernel Two Sample Hypothesis Testing Using the MMD }
\label{sec:background2}
The two sample hypothesis test consists of answering the question of whether we can distinguish between two probability distributions $P$ and $Q$ given  only two 
sets of independent and identically distributed    sets $X=\{{\bf x}_1, ..., {\bf x}_n \}$ and $Y=\{{\bf y}_1, ..., {\bf y}_m \}$ sampled from $P$ and $Q$ respectively. 
The first approach that comes to mind in solving this problem might be to choose a model and estimate the parameters of the model using the given data sets. 
Such traditional  methods generally don't work well in  high dimensional data spaces with limited data because of poor estimation properties in high dimensions. 

The Maximum Mean Discrepancy (MMD) \cite{MMD} is a nonparametric method for dealing with this problem in a Reproducing Kernel Hilbert Space $\cal{H}$ and can be explained as follows. First we  define the mean embedding 
${\bm \mu}_P \in {\cal H}$ of the distribution  $P$ to be the expectation under $P$ of the  mapping $k(.,t)=\Phi(t)$ which can also be written as
\begin{equation}
{\bm \mu}_P(t)=<{\bm \mu}_P(.),k(.,t)>_{\cal H}=E_{\cal X}[k(x,t)].
\end{equation}
The maximum mean discrepancy (MMD)  of the two embedded distributions $P$ and $Q$ is now expressed as the squared difference between
their respective embedded means ${\bm \mu}_P$ and ${\bm \mu}_Q$  as
\begin{eqnarray}
\label{eq:MMDDEf}
MMD(P,Q)=||{\bm \mu}_P - {\bm \mu}_Q||^2_{\cal H}.
\end{eqnarray}
It can be shown that under certain non-restricting conditions on the RKHS,  $MMD(P,Q) = 0$ if and only if $P=Q$. In other words the MMD is injective \cite{MMD}.

Given two independent and identically distributed  sets $X=\{{\bf x}_1, ..., {\bf x}_n \}$ and $Y=\{{\bf y}_1, ..., {\bf y}_m \}$ sampled from $P$ and $Q$  the empirical MMD can be readily derived using the empirical mean embeddings as
\begin{eqnarray}
\label{eq:MMDPrac}
&&\widehat{MMD}(P,Q) = ||\hat{\bm \mu}_P - \hat{\bm \mu}_Q||^2_{\cal H} \\
&=&\frac{1}{n^2}\sum_{i=1}^{n}\sum_{j=1}^{n}K({\bf x}_i{\bf x}_j) -\frac{2}{nm}\sum_{i=1}^{n}\sum_{j=1}^{m}K({\bf x}_i{\bf y}_j) \\
&+& \frac{1}{m^2}\sum_{i=1}^{m}\sum_{j=1}^{m}K({\bf y}_i{\bf y}_j).
\end{eqnarray}
and shown to have favorable performance guarantees. 
Along with the injective property, the empirical MMD can be used to deal with the two sample hypothesis test problem by checking to see if the empirical MMD is
less than a learned threshold. If the MMD of the two sets $X$ and $Y$ is below the threshold and thus sufficiently close to zero then $X$ and $Y$ are likely to have been sampled from the same distribution and we conclude that $P=Q$. This approach makes no model assumptions and is nonparametric and has been shown to have state of the art performance  on high dimensional data sets with limited data \cite{TwoMMD, MMD}.

\section{An RKHS for Sets}
\label{sec:SetRKHS}
In this section we construct an  RKHS for sets of vectors. In later sections we will use this RHKS to deal with the two sample hypothesis test problem in a more effective manner.  
\begin{definition}
\label{def:SetKernelDef}
We define the Set-Kernel on two sets of vectors $X=\{{\bf x}_1, ..., {\bf x}_n \}$ and $Y=\{{\bf y}_1, ..., {\bf y}_m \}$ where  
${\bf x}_i, {\bf y}_i \in \mathbb{R}^n$ as
\begin{eqnarray}
\label{eq:SetKernel}
K(X,Y)=\frac{1}{nm}\sum_{i=1}^{n}\sum_{j=1}^{m}k({\bf x}_i,{\bf y}_j),
\end{eqnarray}
where $k({\bf x}_i,{\bf y}_j)$ is a positive definite kernel with associated mapping $\Phi: \cal{X} \rightarrow \cal{H}$.
Also, the associated Set-Mapping $\Gamma: X \rightarrow  \frac{1}{n}\sum_{i=1}^{n} \Phi({\bf x}_i) \in \cal{H}$ is defined as 
\begin{eqnarray}
\label{eq:SetMapping}
\Gamma(X)=\frac{1}{n}\sum_{i=1}^{n} \Phi({\bf x}_i).
\end{eqnarray}
\end{definition}

\begin{figure}[tbp]   
  \centering
  \includegraphics[width=3.5in]{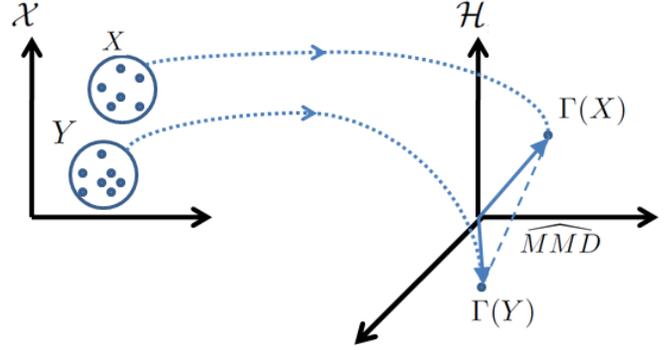}
  \caption{ A visualization of the mapping of sets $X$ and $Y$ into the RKHS associated with the Set-Kernel using the Set-Mapping
	$\Gamma$ and its connection to the empirical MMD as a distance between the vectors $\Gamma(X)$ and $\Gamma(Y)$.}
  \label{fig:KernelSetMap}
\end{figure}

The Set-Kernel is a valid inner product kernel and thus positive definite.
To show this we first write the Set-Kernel in the form of an inner product as
\begin{eqnarray}
K(X,Y)&=&\frac{1}{nm}\sum_{i=1}^{n}\sum_{j=1}^{m}k({\bf x}_i,{\bf y}_j) \\
&=&\frac{1}{nm}\sum_{i=1}^{n}\sum_{j=1}^{m} <\Phi({\bf x}_i),\Phi({\bf y}_j)>_{\cal{H}} \\
&=&<\frac{1}{n}\sum_{i=1}^{n} \Phi({\bf x}_i) , \frac{1}{m}\sum_{j=1}^{m} \Phi({\bf y}_j)>_{\cal{H}} \\
&=&<\Gamma(X), \Gamma(Y)>_{\cal{H}}.
\end{eqnarray}
The positive definiteness of the Set-Kernel is now established by noting that it is the sum of positive definite kernels $k({\bf x}_i,{\bf y}_j)$ and so we write
\begin{eqnarray}
K(X,Y)=\frac{1}{nm}\sum_{i=1}^{n}\sum_{j=1}^{m}k({\bf x}_i,{\bf y}_j) \ge 0. 
\end{eqnarray}
The Set-Kernel is similar to the Derived Subset Kernel defined in \cite{book:KernelMethodsforPatternAnalysis} 
and exactly the same as those proposed in \cite{SetKernelOrigin,SetKernelOrigin2}. 
This specific definition allows for a direct connection to the empirical MMD as follows.  First, we note that the induced norm  in this RKHS for sets can be written as
\begin{eqnarray}
\label{eq:SetNorm}
||\Gamma(X)||_{\cal{H}}^2=<\Gamma(X), \Gamma(X)>_{\cal{H}} \\
= \frac{1}{n^2}\sum_{i=1}^{n}\sum_{j=1}^{n}k({\bf x}_i,{\bf x}_j).
\end{eqnarray}
The empirical MMD is now equal to the distance between two vectors in the RKHS for sets squared. 
Formally,
Let $X=\{{\bf x}_1, ..., {\bf x}_n \}$ and $Y=\{{\bf y}_1, ..., {\bf y}_m \}$, where  
${\bf x}_i, {\bf y}_i \in \mathbb{R}^n$, be two independent and identically distributed sets  sampled from the two respective distributions of $P$ and $Q$.
Also, let $k({\bf x}_i,{\bf y}_j)$ be a positive definite kernel with associated mapping $\Phi: \cal{X} \rightarrow \cal{H}$
and $K(X,Y)$ be the Set-Kernel of (\ref{eq:SetKernel}) with associated Set-Mapping $\Gamma: X \rightarrow  \frac{1}{n}\sum_{i=1}^{n} \Phi({\bf x}_i) \in \cal{H}$,  
then
\begin{eqnarray}
\label{eq:SetDist}
||\Gamma(X)-\Gamma(Y)||_{\cal{H}}^2=\widehat{MMD}(P,Q).
\end{eqnarray} 
We can prove the above equality by writing
\begin{eqnarray}
&&||\Gamma(X)-\Gamma(Y)||_{\cal{H}}^2 \\
&=& <\frac{1}{n}\sum_{i=1}^{n} \Phi({\bf x}_i) - \frac{1}{m}\sum_{i=1}^{m} \Phi({\bf y}_i) , \\
&&\frac{1}{n}\sum_{i=1}^{n} \Phi({\bf x}_i) - \frac{1}{m}\sum_{i=1}^{m} \Phi({\bf y}_i)>_{\cal{H}} \\
&=& \widehat{MMD}(P,Q).
\end{eqnarray}
Figure~\ref{fig:KernelSetMap} visualizes the connections between the empirical MMD and the Set-Kernel mapping.

In summary, we now have an RKHS where the inner product between sets is defined by (\ref{eq:SetKernel}), the norm of a set is defined by (\ref{eq:SetNorm}) and the distance between sets is defined by (\ref{eq:SetDist}). 
Any  kernalized classification algorithm based on computing  inner products, norms or distances can now be applied to sets of vectors. 

In the next  section  we first prove the interesting result that the average probability of error for classifying a set is less than the Bayes error associated with the problem.  We then  specifically solve the two sample hypothesis testing problem by treating it as a one-class classification problem in the proposed RKHS for sets.


\section{The Average Probability of Error for Classifying a Set}
\label{sec:SetRisk}
 A classifier is  a mapping from a feature vector ${\bf x} \in \cal X$ to a class 
label $z \in \{-1,1\}$. Class labels $z$ and feature vectors ${\bf x}$ are sampled from the probability distributions $P_{Z|X}(z|{\bf x})$ and $P_{\bf X}({\bf x})$ respectively. If we only have a single data point ${\bf x}$, the probability of error given ${\bf x}$ is
\begin{eqnarray}
 P(\mbox{error}|{\bf x}) = \left\{ \begin{array}{ll}
         P(z=1|{\bf x}) & \mbox{if we decide $z=-1$};\\
        P(z=-1|{\bf x}) & \mbox{if we decide $z= 1$}.\end{array} \right. 
\end{eqnarray}
The average probability of error for a single point is known as the Bayes Error or Bayes Risk and can be derived as 
\begin{eqnarray}
 &&\mbox{Bayes Error} = \int P(\mbox{error}|{\bf x})P({\bf x}) d{\bf x} = \\
&&\int_{z=-1} P(1|{\bf x})P({\bf x}) d{\bf x} + \int_{z=1} P(-1|{\bf x})P({\bf x}) d{\bf x} = \nonumber \\
&&\frac{1}{2}\int_{z=-1} P({\bf x}|1)d{\bf x} + \frac{1}{2}\int_{z=1} P({\bf x}|-1)d{\bf x}, \nonumber
\end{eqnarray}
where we have assumed equal priors$P(z=1)=P(z=-1)=\frac{1}{2}$. The first term  is known as the miss rate (type-I error)
and the second term is known as the false positive rate (type-II error). It is well known that for fixed data distributions  $P({\bf x}|1)$ 
and $P({\bf x}|-1)$, an average probability of error that is less than the Bayes Error is not possible. Note that this is all conditioned on the assumption that we are basing our decisions on a single data point ${\bf x}$. 

Interestingly, we show that the Bayes Error can be beaten if we base our decisions on a set of data points. 
Specifically, if we have a set of $n$ data points ${\bf x}_1, ..., {\bf x}_n$ that are all identically and independently sampled from the same class, then the probability of error (miss-classifying all $n$ points) given   ${\bf x}_1, ..., {\bf x}_n$ is 
\begin{eqnarray}
 P(\mbox{error}|{\bf x}_1, ..., {\bf x}_n) = \left\{ \begin{array}{ll}
         P(z=1|{\bf x}_1, ..., {\bf x}_n) & \mbox{if  $z=-1$};\\
        P(z=-1|{\bf x}_1, ..., {\bf x}_n) & \mbox{if  $z= 1$},\end{array} \right. \nonumber
\end{eqnarray}
where  $P(z=1|{\bf x}_1, ..., {\bf x}_n)$ is the probability that all $n$ points were sampled from class $z=1$. 

Using the chain rule of probability and the fact that 
${\bf x}_1, ..., {\bf x}_n$  are all sampled independently we can write
\begin{eqnarray}
 P(z|{\bf x}_1, ..., {\bf x}_n) = \frac{P({\bf x}_1|z)... P({\bf x}_n|z)P(z)}{P({\bf x}_1)... P({\bf x}_n)}. 
\end{eqnarray}
We can now derive the average probability of error for a set which we denote as the Set Bayes Error
\begin{eqnarray}
 &&\mbox{Set Bayes Error} = \\
&&\int P(\mbox{error}|{\bf x}_1, ..., {\bf x}_n)P({\bf x}_1, ..., {\bf x}_n) d{\bf x}_1... d{\bf x}_n  = \nonumber\\
&&  \int P(\mbox{error}|{\bf x}_1, ..., {\bf x}_n)P({\bf x}_1)... P({\bf x}_n) d{\bf x}_1... d{\bf x}_n  = \nonumber\\
&& \frac{1}{2}\int_{z=-1}P({\bf x}_1|1)... P({\bf x}_n|1) d{\bf x}_1... d{\bf x}_n + \nonumber \\
&& \frac{1}{2}\int_{z=1}P({\bf x}_1|-1)... P({\bf x}_n|-1) d{\bf x}_1... d{\bf x}_n = \nonumber \\
&& \frac{1}{2}\int_{z=-1}P({\bf x}_1|1)d{\bf x}_1... \int_{z=-1}P({\bf x}_n|1)d{\bf x}_n + \nonumber \\
&& \frac{1}{2}\int_{z=1}P({\bf x}_1|-1)d{\bf x}_1... \int_{z=1}P({\bf x}_n|-1)d{\bf x}_n = \nonumber \\
&& \frac{1}{2}\left( \int_{z=-1}P({\bf x}|1)d{\bf x} \right)^n + 
\frac{1}{2}\left( \int_{z=1}P({\bf x}|-1)d{\bf x} \right)^n. \nonumber 
\end{eqnarray} 
Since $\int_{z=-1}P({\bf x}|1)d{\bf x}$ and $\int_{z=1}P({\bf x}|-1)d{\bf x}$ are  less than or equal to one then
\begin{eqnarray}
&&\frac{1}{2}\left( \int_{z=-1}P({\bf x}|1)d{\bf x} \right)^n + \frac{1}{2}\left( \int_{z=1}P({\bf x}|-1)d{\bf x} \right)^n \le \nonumber\\
&& \frac{1}{2}\int_{z=-1} P({\bf x}|1)d{\bf x} + \frac{1}{2}\int_{z=1} P({\bf x}|-1)d{\bf x}
\end{eqnarray} 
and so 
\begin{eqnarray}
\mbox{Set Bayes Error} \le \mbox{ Bayes Error}.
\end{eqnarray}

The above result is intuitive since it states that the  average probability of error is less if we  base our decision on a set of data samples  rather than a single sample data point. It also states that the average probability of error decrease as a power of $n$ number of sample data points. This important result serves as a motivation for the next section in which we define the hypothesis testing problem as a set classification problem.

\section{The Two Sample Hypothesis Test as  a One-Class Set Classification Problem}
\label{sec:HypothesisAsSetClassify}
In a  two sample hypothesis test problem we are only provided with a set  $X=\{{\bf x}_1, ..., {\bf x}_n \}$ of samples from a distribution $P$. This is the only training data we have and we typically do not get to see any samples from any other alternative distribution. During testing we are presented with a set 
$Y=\{{\bf y}_1, ..., {\bf y}_m \}$ and we must decide if $Y$ consists of
samples from the distribution $P$ or not. 

In view of the RKHS for sets, the two sample hypothesis test problem can now be treated as a one-class classification problem.
Specifically, we need to classify  $Y$ and decide if it belongs to class $P$ or not. The training data consists of  non empty subsets of 
$X=\{{\bf x}_1, ..., {\bf x}_n \}$, such as $X_1=\{{\bf x}_1\}$, $X_2=\{{\bf x}_1, {\bf x}_2 \}$, ..., $X_l=\{{\bf x}_1, ..., {\bf x}_n \}$ which are  each considered a vector  in the RKHS of sets. Any kernalized one-class classification algorithm can learn the reject region in the RKHS of sets and decide if a test vector $Y$ is in the reject region or not. In this paper we specifically use the effective one-class SVM described in \cite{OneClassSVM} which solves the following optimization problem
\begin{eqnarray}
\!\!\!\!\!\!\!\!\!\!\!&&\arg \min_{w,\xi_i,\rho} \frac{1}{2} ||w||^2 + \frac{1}{\upsilon l}\sum_{i=1}^l \xi_i - \rho  \\
\!\!\!\!\!\!\!\!\!\!\! && \mbox{s.t. }  <w, \Gamma(X_i)>_{\cal{H}} ~\geq \rho-\xi_i \\  
 \!\!\!\!\!\!\!\!\!\!\!    && ~~~~~ \xi_i \geq 0.
\end{eqnarray} 
The associated dual problem is 
\begin{eqnarray}
\!\!\!\!\!\!\!\!\!\!\!&&\arg \min_{\alpha_i} \frac{1}{2} \sum_{i=1}^l\sum_{j=1}^l \alpha_i \alpha_j K(X_i,X_j)  \\
\!\!\!\!\!\!\!\!\!\!\! && \mbox{s.t. }   0 \leq \alpha_i \leq \frac{1}{\upsilon l} \\  
 \!\!\!\!\!\!\!\!\!\!\!    && ~~~~~ \sum_{i=1}^l \alpha_i =1.
\end{eqnarray} 
The final decision function applied to a test set $Y$ is  derived from the dual problem and  can be written as 
\begin{eqnarray}
f(Y)=\mbox{sign}\left( \sum_{i=1}^l \alpha_i K(X_i, Y) - \rho \right),
\end{eqnarray}
where $\alpha_i$ are the dual parameters and $K(X_i, Y)$ is the Set-Kernel with associated   Set-Mapping $\Gamma(X_i)$.
Two sample hypothesis testing using   one-class SVM and Set-Kernels is summarized in Algorithm-\ref{alg:ALG1}.

\begin{algorithm}[tb]
{\small
\caption{Kernel Two-Sample Hypothesis Testing Using One-Class SVM and Set-Kernels} 
\label{alg:ALG1}
\begin{algorithmic}
\STATE {\bf Input:} Training samples  $X=\{{\bf x}_1, ..., {\bf x}_n \}$,  Set-Kernel function $K$ and one-class SVM parameter $\upsilon$.
\STATE 1: Make training data $X_1$, $X_2$ ..., $X_l$ composed of nonempty subsets of $X$.
\STATE 2: Solve dual problem 
\begin{eqnarray}
\!\!\!\!\!\!\!\!\!\!\!&&\arg \min_{\alpha_i} \frac{1}{2} \sum_{i=1}^l\sum_{j=1}^l \alpha_i \alpha_j K(X_i,X_j)  \nonumber \\
\!\!\!\!\!\!\!\!\!\!\! && \mbox{s.t. }   0 \leq \alpha_i \leq \frac{1}{\upsilon l} \nonumber \\  
 \!\!\!\!\!\!\!\!\!\!\!    && ~~~~~ \sum_{i=1}^l \alpha_i =1. \nonumber
\end{eqnarray} 
for $\alpha_i$. 
\STATE 3: Find one-class SVM bias parameter $\rho$.
\STATE {\bf Output:} Decision rule $f(Y)=\mbox{sign}\left( \sum_{i=1}^l \alpha_i K(X_i, Y) - \rho \right)$.
\end{algorithmic}
}
\normalsize
\end{algorithm}

While the MMD is a state of the art method for hypothesis testing, its shortcoming is evident when viewed from the standpoint of Set-Kernels. It trains 
by measuring the distances between different data points in the RKHS of sets and learns  a single threshold on these measured distances. If a test 
set has a distance from the training data set that is above the learned threshold then it is rejected. This is a primitive classification method when compared to the 
one-class SVM with Set-Kernels. In the next section we compare the MMD and proposed method on different simulated and real world data sets.

\section{Experimental Results}
\label{sec:Exp}
In this section we perform two sample hypothesis tests on both simulated Gaussian data, 
and benchmark  cancer gene expression data sets and compare the performance of the MMD, F-Test and T-Test methods against the proposed one-class SVM with Set-Kernels method.

\subsection{Simulated Gaussian Data Sets}

Two sets of experiments are reported on simulated Gaussian data. In the first set of experiments we simulated  data from two Gaussians of  equal means and
different variances on a range of different dimensions. We used two Gaussians, $P=N(\bf{0},\sigma_1I)$ and $Q=N(\bf{0},\sigma_2I)$, where we fixed $\sigma_1=1.5$ 
and $\sigma_2=3.5$ and varied the dimensions of the two Gaussians over Dim$=\{2, 5, 10, 25, 50\}$. We sampled $1250$ points from the training distribution $P$ of 
which $250$ were used for training and the other $1000$ were used for testing the type-I error. Also, another  $1000$ points were sampled from $Q$ and used to test the type-II error.  Distinguishing between the two distributions of $P$ and $Q$, that only differ slightly in variance, is generally considered
a challenging problem especially with such limited training samples. 

The MMD reject threshold was found using the standard procedure
from $100$ bootstrap iterations for a fixed type-I error of $\alpha=0.05$. We used the Gaussian embedding as the base kernel $k$ in all our experiments and the Gaussian kernel parameter was found using the median heuristic of \cite{MMD}. The standard two sample  F-test for equal variances was also performed for a fixed type-I error 
of $\alpha=0.05$. Finally, the one-class SVM was trained using the LibSVM toolbox \cite{LIBSVM} with $\upsilon=0.1$ and precomputed Set-Kernels with $100$ random subsets of fixed set size. The base kernel $k$ was a Gaussian with fixed kernel parameter equal to $10$ and the SVM threshold parameter $\rho$
was found from cross validation. Finally, a fixed training and testing set size of $7$ elements each was used for all methods.

\begin{figure}[tbp]   
  \centering
  \includegraphics[width=3.5in]{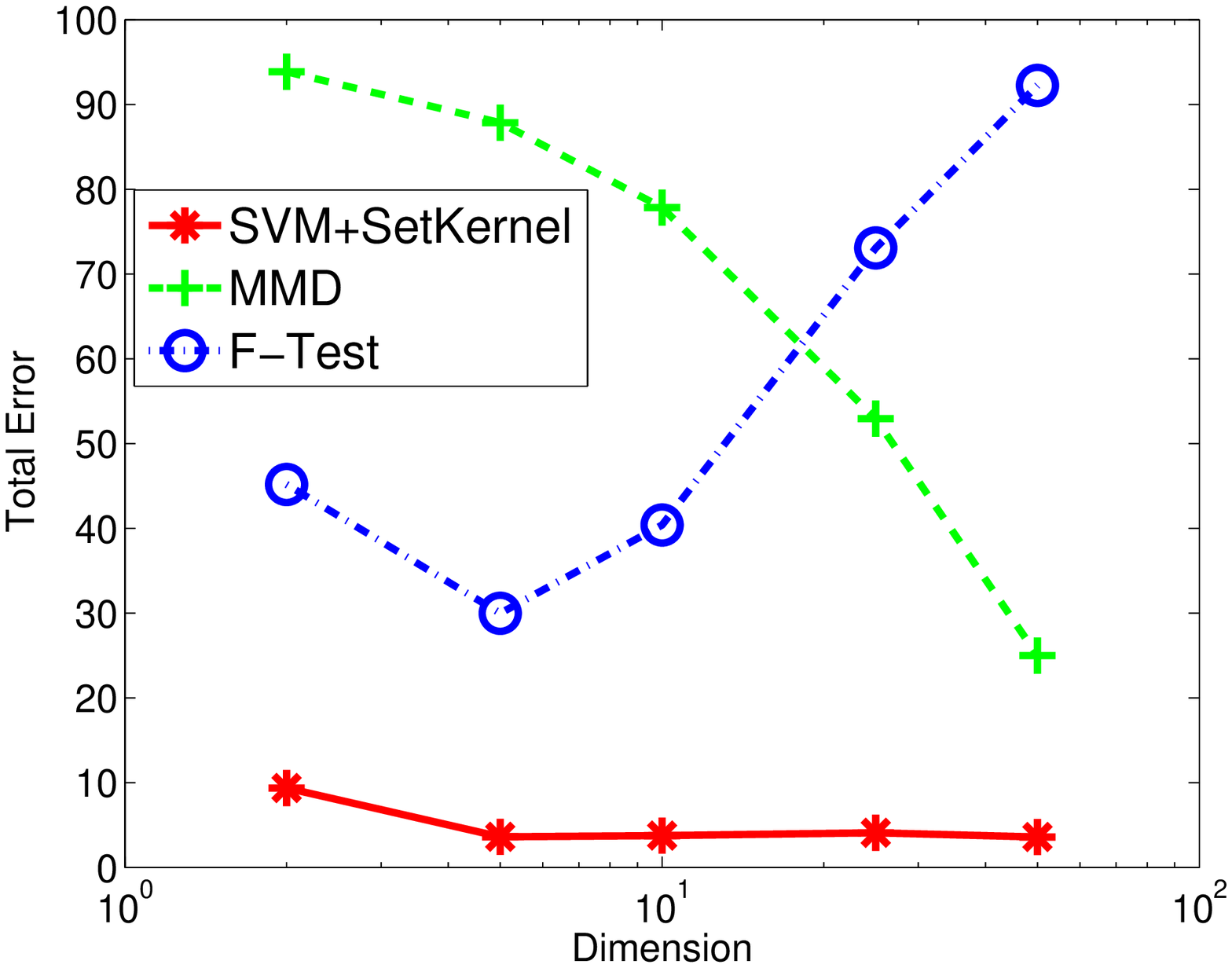}
  \caption{ Total  Error for Gaussian data of different  dimensions using different methods where Total Error = type-I + type-II from Table \ref{tab:Gauss1}. }
  \label{fig:Gauss1}
\end{figure}

The type-I and type-II error test results averaged over $100$ repetitions are reported  in Table \ref{tab:Gauss1} for the MMD, F-Test and one-class SVM with Set-Kernels (SVM+SetKernel) methods. To better visualize the contrast in performance between the three methods we have also plotted the sum of both type-I and type-II errors as Total Error over different dimensions in 
Figure \ref{fig:Gauss1}. 

The one-class SVM with Set Kernels has markedly smaller type-II error at about the same type-I error for all dimensions which results in significant lower Total Error as see in Figure \ref{fig:Gauss1}. While the risk associated with the problem  decreases at higher dimensions, the F-Test does worse at dimensions higher than $d=10$ which emphasizes its inability to deal with high dimensional data. The MMD has the opposite problem of doing poorly on low dimensional data sets \cite{MMDBio}, yet still has markedly higher error than the one-class SVM with Set Kernels even in high dimensions.

\begin{figure}[tbp]   
  \centering
  \includegraphics[width=3.5in]{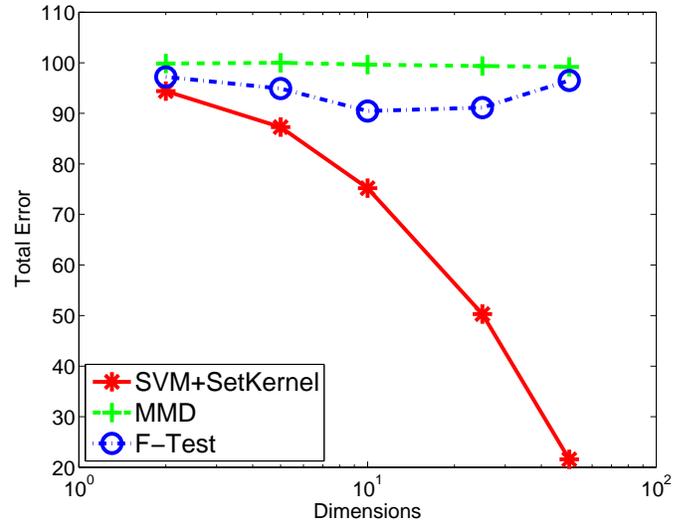}
  \caption{ Total  Error for Gaussian data of different  dimensions using different methods where Total Error = type-I + type-II from Table \ref{tab:Gauss2}.}
  \label{fig:Gauss2}
\end{figure}

In the second set of experiments we repeated the above  under the same conditions  but considered  a much more difficult problem where we fixed $\sigma_1=1.5$ 
and $\sigma_2=1.7$.  
The type-I and Type-II error test results for $100$ repetitions are reported 
in Table \ref{tab:Gauss2} for the MMD, F-Test and one-class SVM with Set-Kernels (SVM+SetKernel) methods. We have  also plotted the sum of both type-I and type-II errors as Total Error over different dimensions in Figure \ref{fig:Gauss2}. 

The one-class SVM with Set Kernels has significatly smaller total error at all dimensions. The F-Test, which is based on estimation techniques, completely breaks down in the high dimensions while the MMD method also performs poorly at all dimensions because it generally has problems in dealing with data sets of equal means and different variances.


\begin{table*}[t]
  \caption{Type-I and type-II error test results on varying low to high dimensional (Dim) Gaussian data ($\sigma_1=1.5$ 
and $\sigma_2=3.5$) averaged over $100$ repetitions  for the MMD, F-Test and one-class SVM with Set-Kernels.} 
  \begin{center}
    \begin{small}
      \begin{tabular}{|c||c|c||c|c||c|c|}
        \hline
        & \multicolumn{2}{c||}{SVM+SetKernel}&\multicolumn{2}{c||}{MMD }&\multicolumn{2}{c||}{F-Test} \\
                                \cline{2-7}
                        &Type-I&Type-II&Type-I&Type-II&Type-I&Type-II \\
                                \hline
        Dim=2 & $4.76\%$ & $4.59\%$ & $5.9\%$ & $87.96\%$ & $10.03\%$ & $35.15\%$   \\
        \hline
        Dim=5     & $3.57\%$ & $0.04\%$ & $5.28\%$ & $82.57\%$ & $21.82\%$ & $8.15\%$   \\
        \hline
        Dim=10   & $3.76\%$ & $0.0\%$ & $6.03\%$ & $71.81\%$ & $39.6\%$ & $0.78\%$   \\
        \hline
        Dim=25    & $4.08\%$ & $0.0\%$ & $5.23\%$ & $47.70\%$ & $73.08\%$ & $0.0\%$   \\
        \hline
        Dim=50     & $3.59\%$ & $0.0\%$ & $5.67\%$ & $19.32\%$ & $92.24\%$ & $0.0\%$   \\
        \hline
      \end{tabular}
    \end{small}
  \end{center}
  \label{tab:Gauss1}
\end{table*}

\begin{table*}[t]
  \caption{Type-I and type-II error test results on varying low to high dimensional (Dim) Gaussian data ($\sigma_1=1.5$ 
and $\sigma_2=1.7$) averaged over $100$ repetitions  for the MMD, F-Test and one-class SVM with Set-Kernels.} 
  \begin{center}
    \begin{small}
      \begin{tabular}{|c||c|c||c|c||c|c|}
        \hline
        & \multicolumn{2}{c||}{SVM+SetKernel}&\multicolumn{2}{c||}{MMD }&\multicolumn{2}{c||}{F-Test} \\
                                \cline{2-7}
                        &Type-I&Type-II&Type-I&Type-II&Type-I&Type-II \\
                                \hline
        Dim=2 & $3.61\%$ & $90.79\%$ & $5.61\%$ & $94.25\%$ & $9.53\%$ & $87.67\%$   \\
        \hline
        Dim=5     & $3.66\%$ & $83.63\%$ & $5.80\%$ & $94.20\%$ & $22.45\%$ & $72.46\%$   \\
        \hline
        Dim=10   & $3.43\%$ & $71.78\%$ & $5.85\%$ & $93.78\%$ & $39.61\%$ & $50.85\%$   \\
        \hline
        Dim=25    & $3.11\%$ & $47.2\%$ & $5.51\%$ & $93.86\%$ & $72.4\%$ & $18.73\%$   \\
        \hline
        Dim=50     & $4.27\%$ & $17.38\%$ & $6.14\%$ & $93.06\%$ & $92.63\%$ & $3.87\%$   \\
        \hline
      \end{tabular}
    \end{small}
  \end{center}
  \label{tab:Gauss2}
\end{table*}

%
%

\subsection{Benchmark Cancer Gene Data Sets}
Next we performed two sample hypothesis tests on six high dimensional benchmark gene expression data sets downloaded from \cite{GeneWebsite}. 
The data sets are challenging because of their small sample size and high dimensions, the details of which are provided in Table \ref{tab:GenDetails}.
The experiments involved splitting the positive samples into a train set and a leave out set for testing the type-I error. The  number of positive train samples, positive leave out samples, test negative samples and fixed set size used in each experiment are also reported in Table \ref{tab:GenDetails}.
The train set was used to learn the MMD reject thresholds from $100$ bootstrap iterations for a fixed type-I error of $\alpha=0.05$.  We again used the Gaussian kernel  and the Gaussian kernel parameter was found using the median heuristic of \cite{MMD}. The standard two sample T-Test was also performed for a fixed type-I error of $\alpha=0.05$. The one-class SVM was trained using the LibSVM toolbox \cite{LIBSVM} with $\upsilon=0.1$ and precomputed Set-Kernels with $100$ random subsets of fixed set size detailed in  Table \ref{tab:GenDetails} for each data set.  The base kernel $k$ was a Gaussian with fixed kernel parameter equal to $1$.

The  type-I and type-II error test results averaged over  $100$ repetitions are reported 
in Table \ref{tab:GenResults} for the MMD, T-Test and one-class SVM with Set-Kernels (SVM+SetKernel) methods.
The one-class SVM with Set-Kernels method has $0\%$ type-I and $0\%$ type-II error on all data sets while the MMD method has considerably  higher error rates on all
data sets. The T-Test method completely fails on these data sets.
This is not surprising when considering the fact that these are  generally  very high dimensional  data sets in the range of $d=2000$, $d=7129$ and $d=12533$ dimensions as seen from Table \ref{tab:GenDetails}. Any method based on estimation techniques, such as the T-Test, will fail in such high dimensions while these type of data sets are easily separable in such high dimensions for a one-class SVM with appropriately chosen Set-Kernels.

\begin{table*}[t]
\caption{\protect\footnotesize{Gene data set details.  }}
\centering
\begin{tabular}{|c|c|c|c|c|c|c|  }
    \hline
    Number & Data Set & \# Train Pos. & \#Leave Out Pos. & \# Test Neg. & \# Set Size & \#Dimensions  \\
    \hline
		\#1& Lung Cancer Women's Hospital & 21 & 10   & 150 & 7 & 12533  \\
    \hline
		\#2& Leukemia & 17 & 8   & 47 & 5 & 7129  \\
    \hline
		\#3& Lymphoma Harvard Outcome & 17 & 9   & 32 & 6 & 7129  \\
    \hline
		\#4& Lymphoma Harvard & 13 & 6   & 58 & 4 & 7129  \\
    \hline
		\#5& Central Nervous System Tumor & 14 & 7   & 39 & 4 & 7129  \\
    \hline
		\#6& Colon Tumor & 15 & 7   & 40 & 4 & 2000  \\
    \hline
\end{tabular}
\label{tab:GenDetails}
\end{table*}

\begin{table*}[t]
  \caption{Two sample hypothesis test results on different benchmark cancer gene data sets using different methods.} 
  \begin{center}
    \begin{small}
      \begin{tabular}{|c||c|c||c|c||c|c|}
        \hline
        & \multicolumn{2}{c||}{SVM+SetKernel}&\multicolumn{2}{c||}{MMD }&\multicolumn{2}{c||}{T-Test} \\
                                \cline{2-7}
                        &Type-I&Type-II&Type-I&Type-II&Type-I&Type-II \\
                                \hline
        Lung Cancer Women's Hospital & $0\%$ & $0\%$ & $18.85\%$ & $0.37\%$ & $100\%$ & $0\%$   \\
        \hline
        Leukemia     & $0\%$ & $0\%$ & $27.31\%$ & $6.37\%$ & $100\%$ & $0\%$   \\
        \hline
        Lymphoma Harvard Outcome    & $0\%$ & $0\%$ & $36.24\%$ & $69.68\%$ & $100\%$ & $0\%$   \\
        \hline
        Lymphoma Harvard    & $0\%$ & $0\%$ & $23.07\%$ & $24.90\%$ & $100\%$ & $0\%$   \\
        \hline
        Central Nervous System Tumor     & $0\%$ & $0\%$ & $19.44\%$ & $77.62\%$ & $100\%$ & $0\%$   \\
        \hline
        Colon Tumor     & $0\%$ & $0\%$ & $18.36\%$ & $33.11\%$ & $100\%$ & $0\%$   \\
        \hline
      \end{tabular}
    \end{small}
  \end{center}
  \label{tab:GenResults}
\end{table*}

\section{Conclusion}
In this work we framed the two sample hypothesis test problem as a one-class classification problem in an appropriate RKHS on sets. 
We showed how to map a set into this RKHS using the provably positive definite Set Kernel. Interestingly,  the  empirical MMD
is the induced norm in this RKHS. Under this view,  the MMD method for hypothesis testing can be seen as placing a simple threshold on 
the distances between training sets. 
We proved that the average probability of error for classifying a set of data samples decreases as a power of the number of samples 
and propose to use the  effective one-class SVM classifier to  perform the hypothesis test. This is made possible by 
the appropriately defined positive definite Set-Kernel. Unlike most traditional hypothesis testing methods such as the F-Test and T-Test, the proposed method is
nonparametric meaning that it does not attempt to estimate the parameters of a probability distribution. This makes the proposed method suitable for
applications with limited high dimensional data. Also, unlike the MMD based method, the proposed method uses the one-class SVM classifier and learns a nonlinear decision surface on the data rather than a single threshold. This gives the proposed method much higher discriminating ability and classification accuracy resulting in lower type-I and type-II errors.  We tested the proposed method on a number of  data sets that were designed to evaluate different challenging scenarios. We first considered   Gaussian data sets of equal mean and different variance with a small number of training samples. This is challenging for the F-Test because of the small size of the training set and it is challenging for the MMD method because the data has equal mean and different variance. The proposed method significantly outperformed both the F-Test and MMD given that the one-class SVM with the Set Kernel can learn complicated decision surfaces with limited data in both low and high dimensions. Finally, we considered six  real world high dimensional gene expression data sets with small sample sizes. 
The T-Test completely failed on these high dimensional data sets  and the MMD had suboptimal performance. On the other hand, The one class SVM with Set Kernels had ideal performance with zero type-I and and zero type-II error on all data sets. 

\bibliographystyle{IEEEtran}
\bibliography{IEEEexample}

\end{document}